\documentclass[conference]{IEEEtran}
\IEEEoverridecommandlockouts

\usepackage[numbers,sort&compress]{natbib}
\usepackage{hyperref}
\hypersetup{
    colorlinks=true,    
    linkcolor=blue,     
    citecolor=blue,     
    filecolor=magenta,  
    urlcolor=cyan,      
}
\newcommand{\figref}[1]{\hyperref[#1]{Fig.~\ref*{#1}}}
\newcommand{\secref}[1]{\hyperref[#1]{section~\ref*{#1}}}
\newcommand{\tabref}[1]{\hyperref[#1]{Table~\ref*{#1}}}

\usepackage{multirow}
\usepackage{booktabs}
\usepackage[flushleft]{threeparttable}
\usepackage{amsmath,amssymb,amsfonts}
\usepackage{algorithmic}
\usepackage{graphicx}
\usepackage{textcomp}
\usepackage{xcolor}
\usepackage{tabularx} 
\def\BibTeX{{\rm B\kern-.05em{\sc i\kern-.025em b}\kern-.08em
    T\kern-.1667em\lower.7ex\hbox{E}\kern-.125emX}}
\begin{document}

\title{UniSync: A Unified Framework for Audio-Visual Synchronization}

\author{
    \IEEEauthorblockN{
        Tao Feng\IEEEauthorrefmark{2}$^{,}$\IEEEauthorrefmark{3}$^{,}$\IEEEauthorrefmark{1}, Yifan Xie\IEEEauthorrefmark{2}$^{,}$\IEEEauthorrefmark{1}, Xun Guan\IEEEauthorrefmark{3}, Jiyuan Song\IEEEauthorrefmark{2}, Zhou Liu\IEEEauthorrefmark{2}, Fei Ma\IEEEauthorrefmark{2}$^{,}$\IEEEauthorrefmark{7} and Fei Yu\IEEEauthorrefmark{2}
    }
    \vspace{0.1cm}
    
    \IEEEauthorblockA{
        \IEEEauthorrefmark{2}Guangdong Laboratory of Artificial Intelligence and Digital Economy (SZ), Shenzhen, China \\
    }
    \IEEEauthorblockA{
        \IEEEauthorrefmark{3}Shenzhen International Graduate School, Tsinghua University, Shenzhen, China \\
    \vspace{0.1cm}
    Email: ft23@mails.tsinghua.edu.cn
    }
    \thanks{$^*$~These authors contributed equally to this work.}
    \thanks{$^{**}$~Corresponding author.}
}

\maketitle

\begin{abstract}
    Precise audio-visual synchronization in speech videos is crucial for content quality and viewer comprehension. 
    Existing methods have made significant strides in addressing this challenge through rule-based approaches and end-to-end learning techniques. However, these methods often rely on limited audio-visual representations and suboptimal learning strategies, potentially constraining their effectiveness in more complex scenarios. 
    To address these limitations, we present UniSync, a novel approach for evaluating audio-visual synchronization using embedding similarities. UniSync offers broad compatibility with various audio representations (e.g., Mel spectrograms, HuBERT) and visual representations (e.g., RGB images, face parsing maps, facial landmarks, 3DMM), effectively handling their significant dimensional differences. 
    We enhance the contrastive learning framework with a margin-based loss component and cross-speaker unsynchronized pairs, improving discriminative capabilities. 
    UniSync outperforms existing methods on standard datasets and demonstrates versatility across diverse audio-visual representations. Its integration into talking face generation frameworks enhances synchronization quality in both natural and AI-generated content.
    
\end{abstract}

\begin{IEEEkeywords}
    video analysis, audio-visual synchronization, representation learning, contrastive learning, talking face generation
\end{IEEEkeywords}

\section{Introduction}
\label{sec:introduction}

    Audio-visual synchronization, particularly lip synchronization, is critical in both real-world scenarios and AI-generated content (AIGC). Research \cite{radiocommunicationrelative} indicates that viewers can detect audio-visual desynchronization within a range of +45 to -125 milliseconds, highlighting the importance of precise synchronization. While various techniques like timecode encoding have been developed \cite{amyes2013film}, methods that rely solely on audio and visual content for synchronization have emerged as more versatile solutions \cite{li2021novel}.

\begin{figure}[t] 
        \centering
        \includegraphics[width=\columnwidth]{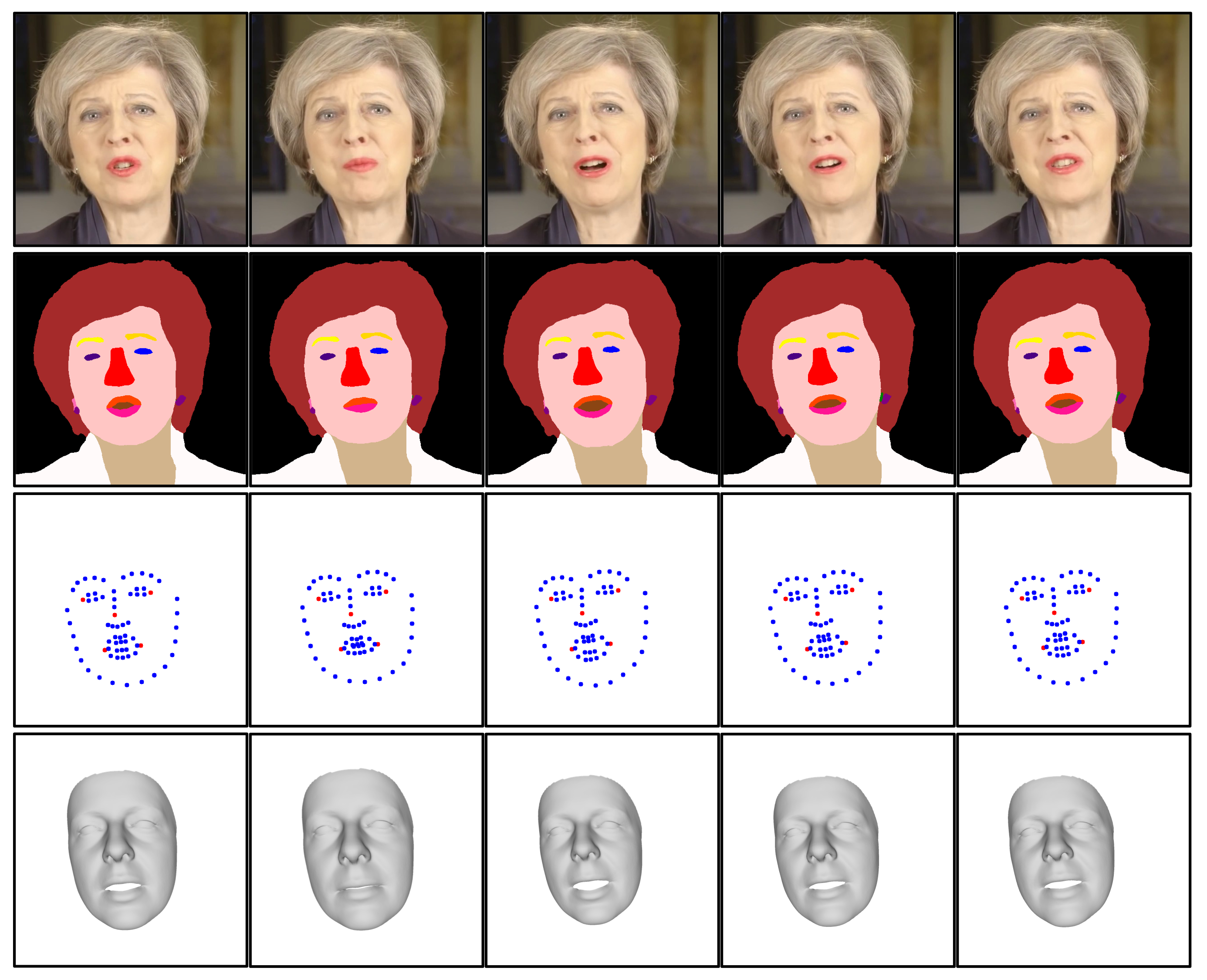}
        \caption{Four most commonly used visual representation methods in speech video analysis. From top to bottom: RGB images, face parsing map, facial landmarks, and 3DMM.}
    \label{fig:four_representations}
    \end{figure}

    However, despite recent advancements in synchronization methods, existing approaches often rely on limited audio-visual representations, failing to leverage diverse representation techniques used in modern video analysis (as shown in \figref{fig:four_representations}). This limitation is particularly evident in talking face generation \cite{ye2023geneface, xiong2024segtalker}, where various audio-visual representations have become essential for enhanced realism (summarized in \tabref{tab:many_talking_face_model}).

    \begin{figure*}[t]
        \centering
        \includegraphics[width=\textwidth]{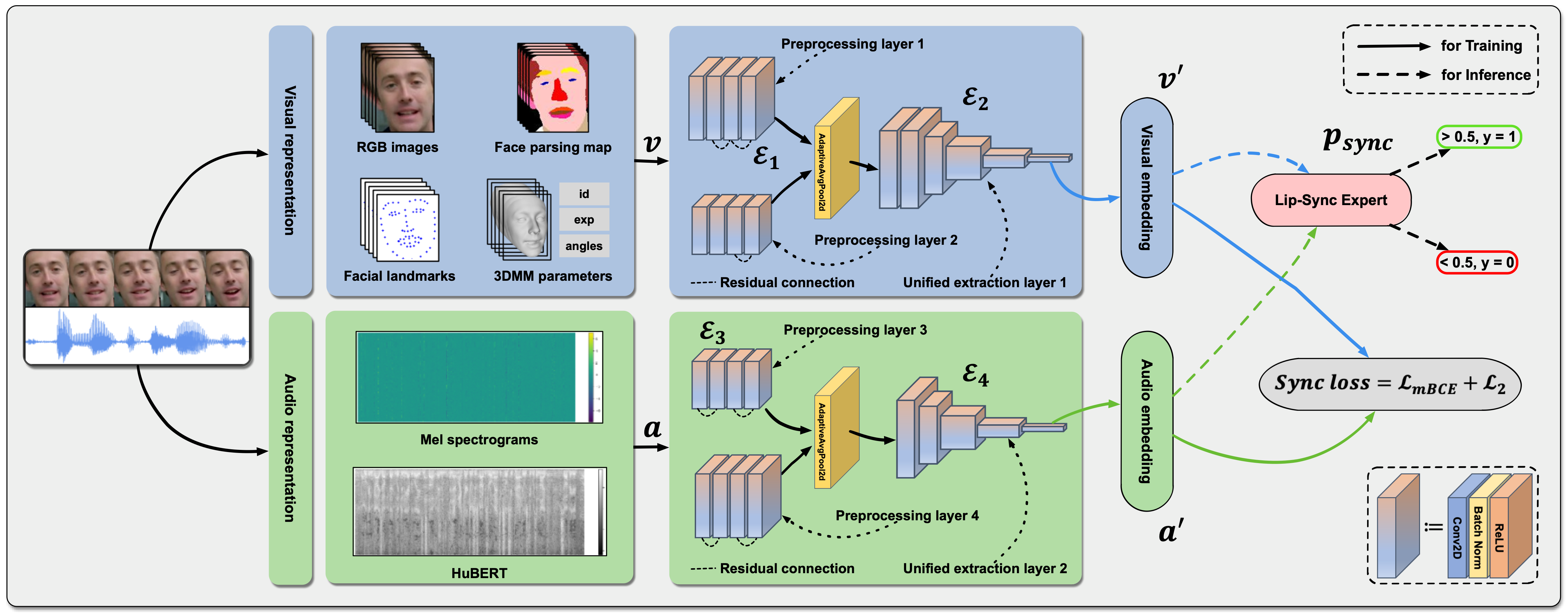}
        \caption{The architecture of UniSync. For a given video segment, the process begins with the extraction of feature vectors using selected visual and audio representation methods, yielding \(v\) and \(a\) respectively. These vectors then undergo initial refinement through specialized preprocessing layers (\(\varepsilon_{1}\) for visual data and \(\varepsilon_{3}\) for audio data). Following an average pooling operation to standardize dimensions, unified feature extraction layers (\(\varepsilon_{2}\) for visual data and \(\varepsilon_{4}\) for audio data) generate the final embeddings \(v'\) and \(a'\). The degree of synchronization between visual and audio content is subsequently determined by computing the cosine similarity (\(p_{\text{sync}}\)) between these refined embeddings.}
    \label{fig:model_structure}
    \end{figure*}

    Furthermore, existing methods often employ inadequate negative sampling strategies \cite{chung2017out, prajwal2020lip, chung2019perfect}, which can hinder their ability to effectively distinguish between synchronized and unsynchronized pairs \cite{kalantidis2020hard}. This becomes particularly problematic when dealing with diverse real-world scenarios or AI-generated talking faces.
    
    To address these challenges, we introduce UniSync, a novel method for evaluating audio-visual synchronization. The main contributions of this study are:
    \begin{itemize}
    \item We propose UniSync, a novel method compatible with multiple audio-visual representation techniques, and investigate the lip synchronization capabilities across different representation combinations.
    \item We enhance the model's performance by introducing a margin-based modification to the binary cross-entropy loss function and incorporating unsynchronized audio-visual pairs from different speakers as negative samples.
    \item We achieve state-of-the-art performance on both LRS2 \cite{afouras2018deep} and CN-CVS \cite{chen2023cn} datasets and conduct comprehensive experiments to demonstrate the versatility and effectiveness of our model.
    \end{itemize}

\section{Related Work}
\subsection{Rule-Based Approaches}
Rule-based approaches rely on predefined linguistic and phonetic rules to map speech sounds to visual representations. These methods typically involve a two-step process: first segmenting speech into phonemes, and then mapping these phonemes to visemes - the visual appearance of sounds on the lips \cite{bear2017phoneme}.

Early works establish important foundations by using linear prediction techniques to decode phonemes from audio and correlate them with precise mouth positions \cite{lewis1991automated}. Subsequent research advances this approach by categorizing English phonemes into 22 distinct visemes and implementing linear interpolation for smoother lip transitions \cite{morishima2002audio}. Further developments propose that the relationship between phonemes and visemes is many-to-many due to visual coarticulation effects, leading to sophisticated mapping schemes using tree-based and K-means clustering techniques \cite{mattheyses2013comprehensive}.

However, these methods often face limitations in real-world applications due to their reliance on controlled environments and extensive phonetic data \cite{wang2010synthesizing, anderson2013expressive, fan2015photo}. These constraints have motivated the development of more flexible, learning-based approaches.

\subsection{End-to-End Learning Methods}
End-to-end learning methods mark a significant advancement by directly learning from raw audio-visual data without relying on intermediate phonetic representations. Early progress emerges with HMM-based techniques that eliminate the need for explicit phoneme labeling \cite{brand1999voice}. The introduction of LSTM networks for direct audio feature processing \cite{shimba2015talking} marks the beginning of deep learning's central role in this field.

A pivotal development comes with the introduction of dual-stream CNN architecture for feature extraction from Mel spectrograms and grayscale images \cite{chung2017out}. This work inspires numerous subsequent developments, including enhanced models with multi-way matching and RGB image processing \cite{chung2019perfect}, cross-modal embedding matrices for accurate synchronization offset prediction \cite{kim2021end}, and multimodal transformer architectures for improved feature fusion \cite{kadandale2022vocalist}.

However, these methods typically rely on limited types of audio-visual representations, primarily using grayscale or RGB images with Mel spectrograms. This constraint overlooks other valuable representation methods that have become crucial in modern video analysis, such as facial landmarks and 3DMM \cite{chen2020comprises}. Our proposed UniSync addresses this limitation by supporting diverse visual and audio representations while maintaining strong performance across different frameworks.

\begin{table}[!t]  
\centering
\caption{Diverse representation techniques used in recent talking face generation methods.}
\label{tab:many_talking_face_model}
\resizebox{\columnwidth}{!}{
\begin{tabular}{lccc}
    \hline
    \textbf{Method} & \textbf{Year} & \textbf{Audio rep.} & \textbf{Visual rep.} \\
    \hline
    Wav2Lip \cite{prajwal2020lip}         & 2020  & Mel spec.   & RGB images    \\
    PC-AVS  \cite{zhou2021pose}           & 2021  & Mel spec.   & RGB images    \\
    AnyoneNet \cite{wang2022anyonenet}    & 2022  & Mel spec.   & Landmarks     \\
    GeneFace \cite{ye2023geneface}        & 2023  & HuBERT      & Landmarks, 3DMM  \\
    DreamTalk \cite{ma2023dreamtalk}      & 2023  & HuBERT      & RGB images, 3DMM \\
    SegTalker \cite{xiong2024segtalker}   & 2024  & Mel spec.   & Face parsing map \\
    PointTalk \cite{xie2024pointtalk}     & 2024  & HuBERT      & RGB images \\
    \hline
\end{tabular}
}
\end{table}

\subsection{Applications in Talking Face Generation}
In the field of talking face generation, lip synchronization plays a crucial role in enhancing content realism. Recent methods employ diverse audio-visual representations, as shown in \figref{fig:four_representations}. Notable examples include Wav2Lip \cite{prajwal2020lip} using Mel spectrograms with RGB images, and GeneFace \cite{ye2023geneface} utilizing HuBERT features with facial landmarks. This diversity in representation methods, summarized in \tabref{tab:many_talking_face_model}, highlights the growing need for versatile synchronization approaches that can accommodate multiple input types.

\section{Methodology}
\label{sec:method}

\subsection{Unified Dual-Stream Architecture}

    UniSync employs a dual-stream architecture to process audio and visual representations separately. The visual stream supports four representation methods: RGB images, face parsing maps, facial landmarks, and 3DMM. These diverse representations capture different aspects of facial motion, from high-level appearance features to precise geometric information. The process involves a preprocessing layer (\(\varepsilon_{1}\)) for initial feature extraction, followed by an adaptive pooling layer to scale features to a fixed size. Larger-scale features like RGB images and face parsing maps undergo dimensionality reduction, while smaller-scale features are subjected to dimensionality increase during preprocessing. Finally, the scaled feature map is processed using a shared extraction layer (\(\varepsilon_{2}\)) to obtain the visual embedding (\(v'\)). 

    The audio stream processes both Mel spectrograms and HuBERT features through parallel pathways. Due to the inherent dimensional differences between these audio representations, we employ specialized preprocessing layers that adapt to their unique characteristics while preserving essential temporal information. Both streams ultimately produce identical-dimension embeddings, enabling direct computation of cosine similarity \( p_{\text{sync}} \). Following established practices \cite{prajwal2020lip}, a threshold of 0.5 serves as the decision boundary between synchronized and unsynchronized pairs.
    
    UniSync's architecture balances simplicity and adaptability, using streamlined neural network components (2D convolutional layers, batch normalization, activation functions, and optional residual connections) to efficiently handle various audio-visual representation methods. The adaptive pooling layers play a crucial role by unifying inputs of different dimensions into a consistent format, enabling the subsequent shared extraction layers to process all representations uniformly. This architectural design maintains computational efficiency while supporting the flexibility needed for multi-modal synchronization tasks.

\begin{figure}[t]  
    \centering
    \includegraphics[width=\columnwidth]{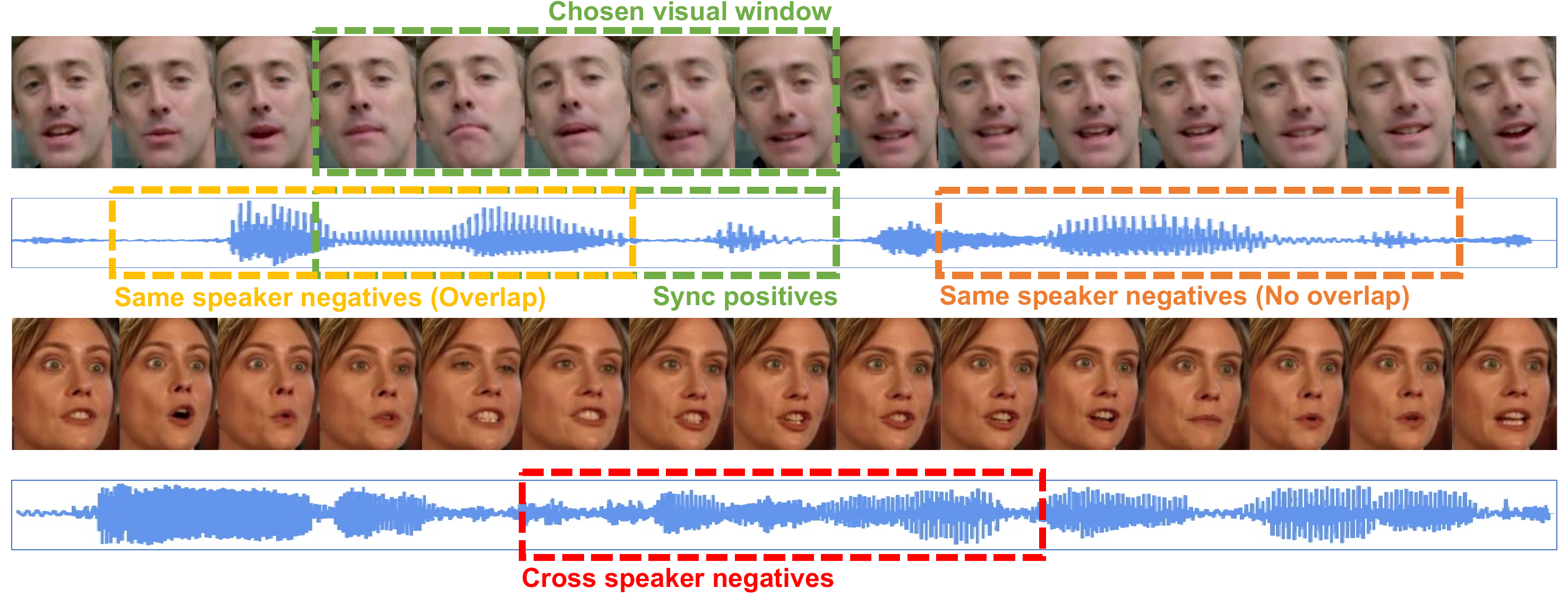}
    \caption{Audio-visual contrastive learning samples: sync positives (matching audio and visual), same speaker negatives (with/without temporal overlap), and cross speaker negatives(mismatched speaker identity).}
    \label{fig:hard_negatives}
\end{figure}

\subsection{Refined Learning Method for Audio-Visual Synchronization}
\label{sec:cross_speaker}

    We introduce a refined contrastive learning approach to enhance audio-visual synchronization detection. Our method combines an innovative loss function with cross-speaker negative samples \cite{oord2018representation}, aiming to maximize cosine similarity for synchronized pairs while minimizing it for unsynchronized ones.
    
\subsubsection{Enhancing with Margin-Based Loss Function}
\label{mBCE_loss}

    To improve the model's ability to discriminate between synchronized and unsynchronized audio-visual pairs, we develop an optimized loss function that combines a margin-based binary cross-entropy loss with \(L_2\) regularization:
    \begin{equation}
    \begin{aligned}
        \mathcal{L}_{\text{ours}} = &-\frac{1}{N} \sum_{i=1}^N \Big[ y_i \log(p_{\text{sync},i}) + \\
        &(1-y_i) \log(1-\max(0,\ p_{\text{sync},i} - m_i)) \Big] + \mathcal{L}_2
    \end{aligned}
    \label{eq:final_loss}
    \end{equation}
    \noindent where \(N\) is the total samples, \(y_i\) the ground truth label, \(p_{sync,i}\) the predicted synchronization probability, and \(m_i\) the margin value. Traditional binary cross-entropy loss may not provide sufficient separation between synchronized and unsynchronized pairs. By introducing distinct margins for positive samples, same-speaker negatives, and cross-speaker negatives, we enforce a more rigorous decision boundary that helps the model better distinguish subtle differences in synchronization quality.
    
    The synchronization probability \(p_{\text{sync}}\) is computed using cosine similarity between audio embedding \(a\) and visual embedding \(v\), each representing a 0.2-second segment:
    \begin{equation}
        p_{\text{sync}} = \cos(a, v) = \frac{a \cdot v}{\Vert a \Vert_2 \cdot \Vert v \Vert_2},
        \label{p_sync}
    \end{equation}
    \noindent where \(\cdot\) denotes the dot product of the vectors, and \(\Vert a \Vert_2\) and \(\Vert v \Vert_2\) are the Euclidean norms of the vectors.
    
    Prior to computing the cosine similarity, both \(a\) and \(v\) undergo a ReLU activation function, ensuring that the calculated \(p_{\text{sync}}\) values range between 0 and 1. This normalization process helps stabilize training and provides interpretable synchronization probabilities, where values closer to 1 indicate higher confidence in synchronization.

    To mitigate overfitting and enhance generalization, we incorporate \(L_2\) regularization:
    \begin{equation}
        \mathcal{L}_{2} = \lambda {\Vert w \Vert}_2^2 = \lambda \sum_{k=1}^K w_k^2
        \label{l2_loss}
    \end{equation}
    \noindent where \(\lambda\) is a regularization parameter that controls the penalty strength, \(K\) is the number of weights in the network, and \(w_k\) represents each individual weight. This regularization term is particularly important given the diverse nature of our audio-visual representations and the need to maintain robust performance across different domains.

\begin{table}[!t]
    \centering
    \caption{Comparison with recent works. Best performance in \textbf{bold}.}
    \label{tab:model_comparison}
    \resizebox{\columnwidth}{!}{%
    \begin{tabular}{clllcc}
    \hline
    \textbf{Year} & \textbf{Models} & \textbf{Backbone} & \textbf{Audio rep.}  &\textbf{Acc.(\%)} \\ 
    \hline
    2016 & SyncNet \cite{chung2017out}    & CNN  & MFCC   & 75.8     \\
    2018 & PM \cite{chung2019perfect}     & CNN  & Mel spec.        & 88.1     \\
    2020 & Wav2Lip \cite{prajwal2020lip}  & CNN  & Mel spec.        & 90.7     \\
    2021 & AVST \cite{chen2021audio}      & Transformer  & Mel spec.    & 92.0     \\
    2022 & VocaLiST \cite{kadandale2022vocalist}   & Transformer & Mel spec.   & \textbf{92.8}  \\
    \midrule
    2024 & UniSync (Ours)              & CNN  & Mel spec.      & 91.77    \\
    2024 & UniSync (Ours)     & CNN  & HuBERT     & \textbf{94.27}    \\
    \hline
    \end{tabular}
    }
\end{table}

\begin{table}[!t]
\centering
\caption{Comparison on Model Efficiency. Best performance in \textbf{bold}.}
\label{tab:efficiency_comparison}
\resizebox{\columnwidth}{!}{%
    \setlength{\tabcolsep}{2.5pt}
    \begin{tabular}{lccccc}
    \hline
    \textbf{Model} & \textbf{FLOPs} & \textbf{Params} & \textbf{Speed*} & \textbf{Conv. Speed**} & \textbf{Acc. (\%)} \\
    \hline
        Wav2Lip~\cite{prajwal2020lip} & 1.22G & 16.4M & 30s & Fast (30) & 90.7 \\
        Vocalist~\cite{kadandale2022vocalist} & 20.17G & 80.1M & 4.25min & Slow (289) & 92.8 \\
        UniSync (Ours) & 1.73G & 16.3M & 33s & Fast (26) & \textbf{94.27} \\
    \hline
    \multicolumn{6}{l}{\footnotesize *Training time per epoch on RTX 4090 GPU} \\
    \multicolumn{6}{l}{\footnotesize **Convergence speed to 80\% val accuracy (epochs needed)} \\
    
    \end{tabular}
    }
\end{table}

\subsubsection{Improving with Cross-Speaker Negative Samples}
\label{negatives_samples}
    
    To further enhance our proposed framework, we focus on effective sample selection, which plays a crucial role in model performance \cite{kalantidis2020hard}. Traditional synchronization methods typically construct negative samples by shifting the audio or visual content within a single speaker's video. However, this approach may be insufficient for modern audio-visual tasks, particularly in talking face generation \cite{ye2023geneface, xiong2024segtalker}, where the fundamental goal is to synchronize target audio with different speakers' facial movements.
    
    Drawing inspiration from previous research \cite{korbar2018cooperative}, we distinguish between two types of negative samples: those from the same speaker and those from different speakers (see \figref{fig:hard_negatives}). By incorporating cross-speaker negative samples, where audio from one speaker is paired with visuals from another, we create a more challenging and realistic training scenario. The effectiveness of this strategy is particularly evident when applying UniSync to talking face generation tasks, where it significantly improves the synchronization quality of generated content (as demonstrated in \tabref{tab:quantitative_results}). Our implementation maintains an appropriate proportion of cross-speaker negative samples in the training process, enabling the model to better handle the diverse synchronization scenarios encountered in real-world applications.

\section{Experiments}
\label{sec:experiments}

\subsection{Datasets and Implementation Details}
\label{subsec:datasets}
    To develop a unified framework compatible with various representation methods and generalizable across diverse speech videos, we utilize two large-scale lip-reading corpora: LRS2 \cite{afouras2018deep} and CN-CVS \cite{chen2023cn}.
    LRS2, sourced from BBC broadcasts, contains over 224 hours of English spoken content from 1,000+ speakers. CN-CVS, compiled from various Chinese sources, comprises 273 hours of Mandarin visual-speech from 2,500+ speakers. Both datasets offer diverse content for lip-sync tasks, encompassing different speaking styles, emotional expressions, and recording environments.

    Our UniSync model is trained using a batch size of 64 on an NVIDIA RTX 4090 GPU with 24GB of memory. The training process employs the Adam optimizer \cite{kingma2014adam} with a learning rate of \(1\times10^{-4}\) and momentum parameters \(\beta_1=0.5\), \(\beta_2=0.999\). To mitigate overfitting, we incorporate weight decay with a coefficient of \(1\times10^{-4}\).

\begin{table}[!t]
\centering
\caption{Lip-sync accuracy for different audio-visual representation combinations on LRS2 dataset. Best performance in \textbf{bold}.}
\label{tab:representations_compare}
\resizebox{\columnwidth}{!}{%
\setlength{\tabcolsep}{2.5pt}
    \begin{tabular}{llcc}
        \hline
        \textbf{Audio rep.} & \textbf{Visual rep.} & \textbf{Max Acc.(\%)} & \textbf{Average Acc.(\%)} \\
        \hline
        HuBERT        & RGB images        & \textbf{93.25} & \textbf{90.09} \\
        HuBERT        & 3DMM              & 91.40 & 89.11 \\
        Mel spec.     & RGB images        & 91.31 & 87.23 \\
        Mel spec.     & 3DMM              & 89.00 & 85.93 \\
        HuBERT        & Landmarks         & 89.00 & 85.26 \\
        HuBERT        & Face parsing map  & 86.04 & 82.50 \\
        Mel spec.     & Landmarks         & 86.23 & 82.39 \\
        Mel spec.     & Face parsing map  & 83.64 & 79.72 \\
        \hline
    \end{tabular}
}
\end{table}

\subsection{Baselines and Evaluation Metrics}
\label{subsec:evaluation_protocol}
    For comprehensive evaluation, we compare our UniSync against several methods: SyncNet \cite{chung2017out}, which pioneers the CNN-based approach with MFCC features; Perfect Match (PM) \cite{chung2019perfect} and Wav2Lip \cite{prajwal2020lip}, which advance the field using Mel spectrograms; and more recent Transformer-based models like AVST \cite{chen2021audio} and VocaLiST \cite{kadandale2022vocalist}.
    
    We employ two primary evaluation metrics: Lip-Sync Accuracy and Lip Sync Error, following the methodology proposed by \cite{prajwal2020lip}. Lip-Sync Accuracy is calculated using randomly selected synchronized and unsynchronized audio-visual pairs. Additionally, we utilize LSE-D (Lip Sync Error - Distance) and LSE-C (Lip Sync Error - Confidence), based on the pre-trained SyncNet model \cite{chung2017out}, to evaluate synchronization accuracy in uncontrolled environments. These industry-standard metrics enable direct performance comparisons with existing and future methods.

\subsection{Comparison of Lip-Sync Accuracy with Popular Models}
\label{sec:lip_sync_acc}

    As shown in \tabref{tab:model_comparison}, experimental results demonstrate the effectiveness of the proposed model, particularly when using HuBERT as audio representation. Our CNN-based UniSync achieves an impressive accuracy of 94.27\%, surpassing the performance of recent transformer-based models such as AVST \citep{chen2021audio} and VocaLiST \citep{kadandale2022vocalist}. This result underscores the potential of combining traditional CNN architectures with state-of-the-art audio processing techniques.
    To ensure a fair comparison with previous methods, we also evaluate UniSync using Mel spectrograms as the audio input. In this configuration, UniSync achieves an accuracy of 91.77\%, which, while slightly lower than the transformer-based methods, still represents a significant improvement over earlier CNN-based approaches like PM \citep{chung2019perfect} and Wav2Lip \citep{prajwal2020lip}.
    
    Furthermore, as illustrated in \tabref{tab:efficiency_comparison}, UniSync demonstrates superior computational efficiency compared to transformer-based alternatives. Our model maintains comparable computational complexity to Wav2Lip \cite{prajwal2020lip} while significantly outperforming VocaLiST \cite{kadandale2022vocalist} in terms of both parameter count and FLOPs. The training efficiency is especially significant, with UniSync requiring substantially less time per epoch and fewer epochs to achieve convergence compared to VocaLiST \cite{kadandale2022vocalist}.

    These results demonstrate the effectiveness of our approach. To further understand the impact of different representation choices, we conduct a comprehensive analysis of various audio-visual combinations in the following section.

\begin{table}[!t]
    \centering
    \caption{Ablation study results. Best performance in \textbf{bold}.}
    \label{tab:ablation_study}
    \begin{threeparttable}
    \setlength{\tabcolsep}{2.5pt}
    \begin{tabular}{lccc}
        \hline
        \textbf{Aspect} & \textbf{Value} & \textbf{Acc.(Mel)} & \textbf{Acc.(HuBERT)} \\
        \hline
        \multirow{3}{*}{Cross-speaker proportion\tnote{1}} & 0.0 & 88.96 & 91.79 \\
                                                 & \textbf{0.2} & \textbf{89.23} & \textbf{92.18} \\
                                                 & 0.5 & 88.93 & 91.96 \\
    
        \hline
        \multirow{3}{*}{Margin (same, cross)\tnote{2}} & (0, 0) & 91.22 & 93.62 \\
                                             & (0.1, 0.3) & \textbf{91.77} & 93.99 \\
                                             & (0.3, 0.7) & 91.50 & \textbf{94.27} \\
        \hline
    \end{tabular}
    \begin{tablenotes}
      \item[1] Results reported as mean accuracy across multiple well-converged runs, ensuring a reliable measure of overall performance.
      \item[2] Results indicate maximum accuracy achieved, demonstrating the optimal potential of each configuration.
    \end{tablenotes}
    \end{threeparttable}

\end{table}

\subsection{Evaluating Lip-Sync Accuracy Across Audio-Visual Representations}
\label{sec:performance_different_representations}

    We evaluate the performance of various audio-visual representation combinations, as shown in \tabref{tab:representations_compare}. The results reveal several important patterns: (1) HuBERT consistently outperforms Mel spectrograms across all visual representations, with improvements ranging from 2.4\% to 3.5\% in average accuracy; (2) Among visual representations, RGB images achieve the best performance (93.25\%), followed by 3DMM (91.40\%), while face parsing maps show relatively lower accuracy; (3) The combination of HuBERT and RGB images yields both the highest maximum accuracy (93.25\%) and average accuracy (90.09\%), suggesting this pairing as the optimal choice for practical applications. Notably, the performance gap between different visual representations is more pronounced when using Mel spectrograms compared to HuBERT, indicating HuBERT's stronger capability in extracting discriminative audio features.

\subsection{Ablation Study}
\label{sec:ablation_study}

    We conduct ablation studies on two critical components of UniSync: the proportion of cross-speaker negative samples and the margin values in the loss function. \tabref{tab:ablation_study} presents our findings using both Mel spectrogram and HuBERT audio representations.
    
    For cross-speaker negative samples, we observe that a moderate proportion of 0.2 achieves optimal performance, improving accuracy from 88.96\% to 89.23\% with Mel spectrograms and from 91.79\% to 92.18\% with HuBERT. Higher proportions (0.5) show diminishing returns, suggesting that a balanced mix of same-speaker and cross-speaker negatives is crucial for effective training.
    
    The introduction of margin values in the loss function consistently improves performance compared to the baseline without margins. As shown in \tabref{tab:ablation_study}, both audio representations benefit from this enhancement, with HuBERT achieving slightly better results overall.
        
\subsection{Improving Lip-Sync Quality in Talking Face Generation}
\label{sec:phase2}

    We integrated UniSync into two representative talking face generation models: Wav2Lip~\cite{prajwal2020lip} and GeneFace~\cite{ye2023geneface}. As shown in \tabref{tab:quantitative_results}, UniSync significantly improves lip synchronization quality across different scenarios. For Wav2Lip, our integration reduces LSE-D from 7.521 to 6.647 on LRS2 (English) and from 11.025 to 9.898 on CN-CVS (Mandarin), while simultaneously increasing LSE-C scores. Similar improvements are observed with GeneFace, where LSE-D decreases from 9.809 to 9.396 and LSE-C increases from 4.765 to 5.294 on real-world video content.
    
    These consistent improvements across different models, languages, and datasets demonstrate UniSync's effectiveness as a general-purpose lip synchronization enhancement module. Notably, the performance gains in both controlled (LRS2) and real-world scenarios (GeneFace's dataset) suggest UniSync's strong adaptability to various practical applications.
    
\begin{table}[!t]
    \centering
    \caption{Quantitative results on LRS2, CN-CVS, and GeneFace's datasets. Best performance in \textbf{bold}.}
    \label{tab:quantitative_results}
    \resizebox{\columnwidth}{!}{
    \setlength{\tabcolsep}{2.5pt}
    \begin{threeparttable}
        \begin{tabular}{lcccccc}
            \hline
            \multirow{2}{*}{\textbf{Method}} & \multicolumn{2}{c}{\textbf{LRS2}} & \multicolumn{2}{c}{\textbf{CN-CVS}} & \multicolumn{2}{c}{\textbf{GeneFace's dataset}\tnote{1}} \\
            \cmidrule(lr){2-3} \cmidrule(lr){4-5} \cmidrule(lr){6-7}
            & \textbf{LSE-D} $\downarrow$ & \textbf{LSE-C} $\uparrow$ & \textbf{LSE-D} $\downarrow$ & \textbf{LSE-C} $\uparrow$ & \textbf{LSE-D} $\downarrow$ & \textbf{LSE-C} $\uparrow$ \\
            \hline
            LipGAN \cite{kr2019towards} & 10.330 & 3.199 & \multicolumn{2}{c}{-} & \multicolumn{2}{c}{-} \\
            Wav2Lip \cite{prajwal2020lip} & 7.521 & 6.406 & 11.025 & 1.439 & \multicolumn{2}{c}{-} \\
            \textbf{Wav2Lip + UniSync} & \textbf{6.647} & \textbf{6.842} & \textbf{9.898} & \textbf{2.2512} & \multicolumn{2}{c}{-} \\
            \hline
            GeneFace \cite{ye2023geneface} & \multicolumn{2}{c}{-} & \multicolumn{2}{c}{-} & 9.809 & 4.765 \\
            \textbf{GeneFace + UniSync} & \multicolumn{2}{c}{-} & \multicolumn{2}{c}{-} & \textbf{9.396} & \textbf{5.294} \\
            \hline
        \end{tabular}
        \begin{tablenotes}
            \item[1] GeneFace's dataset refers to a video of a female speaker approximately 4 minutes long, some frames of which are shown in \figref{fig:four_representations}.
        \end{tablenotes}
    \end{threeparttable}
    }

\end{table}

\section{Conclusion}
\label{sec:conclusion}

We present UniSync, a novel method for audio-visual synchronization that addresses key limitations of existing approaches. UniSync's broad compatibility with various representation techniques, including RGB images, face parsing maps, facial landmarks, and 3DMM for visual content, alongside Mel spectrograms and HuBERT features for audio, enhances its adaptability. Our refined contrastive learning strategy, incorporating both same-speaker and cross-speaker negative samples, substantially improves synchronization evaluation capability.
Through extensive experiments, UniSync demonstrates superior performance in standard benchmarks and enhances lip synchronization quality when integrated into talking face generation models. The method's ability to handle diverse representations while maintaining computational efficiency makes it valuable for real-world applications and future audio-visual tasks, especially in the field of AI-generated content.

\bibliographystyle{IEEEbib}
\bibliography{icme2025_template}

\end{document}